\documentclass[letterpaper]{article} 
\usepackage{aaai24}  
\usepackage{times}  
\usepackage{helvet}  
\usepackage{courier}  
\usepackage[hyphens]{url}  
\usepackage{graphicx} 
\usepackage{natbib}  
\usepackage{caption} 
\usepackage{algorithm}
\usepackage{algorithmic}
\usepackage{bm}
\usepackage{amsfonts}
\usepackage{amsmath}
\usepackage{amssymb}
\usepackage{makecell}
\usepackage{graphicx}
\usepackage{float}
\usepackage{subfig}
\usepackage{multirow}

\usepackage{newfloat}
\usepackage{listings}
\DeclareCaptionStyle{ruled}{labelfont=normalfont,labelsep=colon,strut=off} 
\floatstyle{ruled}
\newfloat{listing}{tb}{lst}{}
\floatname{listing}{Listing}
\nocopyright 

\title{IMM: An Imitative Reinforcement Learning Approach with Predictive Representation Learning for Automatic Market Making}

\author{
    Hui Niu\textsuperscript{\rm 1}\equalcontrib, 
    Siyuan Li\textsuperscript{\rm 3}\equalcontrib, 
    Jiahao Zheng\textsuperscript{\rm 2}, 
    Zhouchi Lin\textsuperscript{\rm 2}, 
    Jian Li\textsuperscript{\rm 1}, 
    Jian Guo\textsuperscript{\rm 2}, 
    Bo An\textsuperscript{\rm 4}
}
\affiliations{
    \textsuperscript{\rm 1}Tsinghua University \qquad\qquad  
    \textsuperscript{\rm 2}International Digital Economy Academy  \\
    \textsuperscript{\rm 3}Harbin Institute of Technology \quad 
    \textsuperscript{\rm 4}Nanyang Technological University\\
    
}

\usepackage{bibentry}

\begin{document}

\maketitle

\begin{abstract}
 Market making (MM) has attracted significant attention in financial trading owing to its essential function in ensuring market liquidity.
 With strong capabilities in sequential decision-making,  Reinforcement Learning (RL)  technology has achieved remarkable success in quantitative trading. 
 Nonetheless, most existing RL-based MM methods focus on optimizing single-price level strategies which fail at frequent order cancellations and loss of queue priority. 
 Strategies involving multiple price levels align better with actual trading scenarios.
 However, given the complexity that multi-price level strategies involves a comprehensive trading action space, the challenge of effectively training profitable RL agents for MM  persists.
 Inspired by the efficient workflow of professional human market makers, we propose Imitative Market Maker (IMM), 
 a novel RL framework leveraging  both knowledge from suboptimal signal-based experts and direct policy interactions to  develop  multi-price level MM strategies efficiently.
 The framework start with introducing effective state and action representations adept at encoding information about multi-price level orders.
 Furthermore,  IMM integrates a representation learning unit capable of capturing both short- and long-term market trends to mitigate adverse selection risk.
 Subsequently, IMM formulates an expert strategy based on signals and trains the agent through the integration of  RL and imitation learning techniques, leading to efficient learning.
 Extensive experimental results on four real-world market datasets demonstrate that IMM outperforms current RL-based market making strategies in terms of several financial criteria. 
 The findings of the ablation study substantiate the effectiveness of the model components.

\end{abstract}


\section{Introduction}

Market making (MM) is a process where a market maker continuously places buy and sell orders on both sides of the limit order book (LOB) of a given security. 
During this process, market makers encounter various risks, including inventory risk, adverse selection risk, and non-execution risk, rendering MM a complex trading task.
Optimal MM entails dynamic adjustment of bids and asks in response to the market maker’s  inventory level and current market status to maximize the risk-adjusted returns.

In contrast to  traditional MM approaches \cite{AS2008,Guant2012}  which rely on mathematical models with strong assumptions, 
(deep) Reinforcement Learning (RL) has emerged as a promising approach for developing MM strategies capable of adapting to changing market dynamics.
While there has been extensive research on the application of RL for MM, the majority of studies  have focused on optimising single-price level policies \cite{Spooner2018MarketMV,Sadighian2019DeepRL,Gueant2019DeepRL,Xu2022PerformanceOD}. 
 Unfortunately, it is pointed out that such strategies result in  frequent and unnecessary order cancellations, leading to the loss of order priority and non-execution risks \cite{ICAIF2022os,ICAIF2022beta}.  
Consequently, strategies enabling traders to place multi-price level orders in advance to reserve good queue positions beyond best bid/ask levels are better suited for realistic MM scenarios.
However, given that multi-price level strategies involves a large fine-grained trading action space than the single-price level ones, how to effectively train profitable RL agents for MM   remains a challenging problem.
Furthermore, the trade-off between profits and the various sources of risk based on personalized risk preferences has not been well addressed.
 To address these challenges, this paper proposes the Imitative Market Maker (IMM), an RL framework that integrates proficient representation learning and imitation learning techniques, to resolve the optimal MM problem using multi-price level policies.

\begin{figure}[t]
  \centering
  \includegraphics[width=\linewidth]{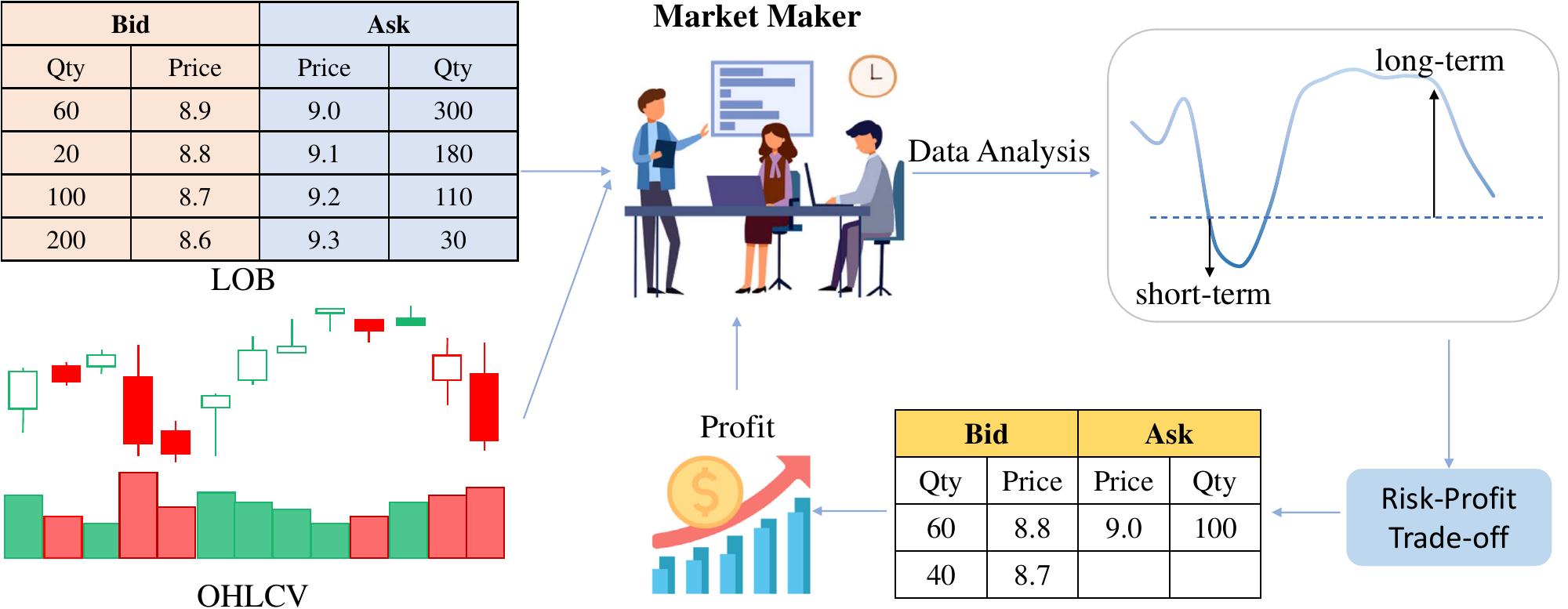}\\
  \caption{Workflow of professional human market makers.}
  \label{fig:workflow}
\end{figure}

The insight of IMM comes from the following inspiration:  
considering the workflow of a skilled human market maker (Figure \ref{fig:workflow}), the professional first gathers both micro- and macro-level market information. The former aids in assessing market liquidity, while the latter contributes to evaluating adverse selection risks.
Subsequently, he/she predicts short-term and long-term market trends based on these market information.
Afterwards, he/she trade off between  risks and profits according to their risk preference, leading to the ultimate trading decision 
 ( the multiple prices and volumes of the quotes).
Within numerous successful trading firms, this workflow plays a pivotal role in deriving robust and profitable MM strategies.
Furthermore, beyond learning from interactions with the environment, extracting trading knowledge from experts presents an appealing method for mining profitable patterns \cite{Ding2018imitation4}.
Leveraging  expert data is a favorable  approach to improve the exploration ability of RL algorithms\cite{Sun2018imitation3,Mendonca2019imitation1}.

 Motivated by these inspirations, IMM integrates a state representation learning unit (SRLU) with an imitative  RL unit (IRLU) to facilitate efficient policy learning for MM.
 Our primary contributions can be summarized as follows:
 \begin{itemize}

   \item 
   IMM introduces effective state and action representations adeptly encoding multi-price level order information.
   These representations are well-suited for the MM environment, enabling the implementation of order stacking.
   Furthermore, IMM incorporates multi-granularity predictive signals as auxiliary variables, and employs a  temporal convolution and spatial attention (TCSA) network to distill  valuable representations from noisy market data. 

    \item 
    IMM utilizes trading knowledge from experts to facilitate effective exploration in the complex trading environment.
    To meet various risk preferences, IRLU trains customized agents by adjusting utility parameters of a specially crafted reward function.

    \item 
    Through extensive experiments on four real-world  financial futures datasets, we demonstrate that IMM significantly outperforms many baseline methods in relation to both risk-adjusted returns and adverse selection ratios.
    We underscore IMM’s practical applicability to MM through a series of comprehensive exploratory and ablative studies.

 \end{itemize}

\section{Related Work}

\subsubsection{Traditional Finance Methods}
Traditional approaches for MM in the finance literature \cite{Amihud1980DealershipMM,Glosten1985BidAA,AS2008,Guant2012}  consider MM as a stochastic optimal control problem that can be solved analytically. 
For example, the Avellaneda–Stoikov (AS) model \cite{AS2008} assumes a drift-less diffusion process for the mid-price evolution, then uses the Hamilton-Jacobi-Bellman equations to derive closed-form approximations to the optimal quotes. 
On a related note, \cite{Guant2012} consider a variant of the AS model with inventory limits.
However, such methods are typically predicated on a set of strong assumptions, and employ multiple parameters that need to be laboriously calibrated on historical data. 
It seems promising to consider more advanced methods such as RL  that enables learning directly from data in a model-free fashion.

\subsubsection{RL-based Methods}

Recent years have witnessed a strong popularity of (deep) RL in the field of quantitative trading \cite{Chan2001AnEM,Cartea2015EnhancingTS,Patel2018OptimizingMM,Zhong2020DataDrivenMV,Fang2021UniversalTF,Niu2022MetaTraderAR,Anbo2023alphamix}.

The majority of the RL-based MM approaches adopts a single-price level strategy.
Among those studies, several methods proposed defining the action space in advance \cite{Spooner2018MarketMV,Xu2022PerformanceOD,Sadighian2019DeepRL}. 
Several researchers utilized a "half-spread" action space which chooses a continuous half spread on each side of the book \cite{Glosten1985BidAA,Jumadinova2010ACO,Cartea2015AlgorithmicAH,Lim2018ReinforcementLF,Gueant2019DeepRL}.  
Unfortunately, when actually implementing such a strategy, to change the half spread on each side of the book, it is necessary to actively cancel orders at each time step and place new orders at the new level. This results in  frequent unnecessary order cancellations, leading to queue position losing \cite{ICAIF2022beta}. 

To overcome this limitation, ladder strategies, which place a unit of volume at all prices in two price intervals, one on each side of the book, has been adopted\cite{Chakraborty2011MarketMA, Abernethy2013AdaptiveMM}. 
The latest work uses variants of this strategy to construct multi-price level MM policies.
The $DRL_{OS}$ \cite{ICAIF2022os} agent decides whether to retain one unit of volume on each price level, and the beta policy \cite{ICAIF2022beta} allows for flexibility in the distribution of order volumes.

However,  to provide more accurate trading decisions, multi-price level strategy involves a much complex fine-grained action space (multi-level price and quantity).
Little attention has been paid to inefficient  exploration problem due to the complex action space of multi-price strategies.
While have shown great promise, these approaches might be limited to achieve efficient exploration,  particularly in highly dynamic and complex market environments.

\section{Imitative Market Maker (IMM)}

 This section introduces the proposed IMM framework.
 We start with introducing a novel state/action space and illustrating the transition dynamics of MM procedure that accommodates multi-price level order stacking.  Subsequently, we elaborate on the SRLU which aims at forecasting multi-granularity signals while extracting valuable representations from the noisy market data.  Lastly, we outline the MM policy learning approach,  which incorporates the RL and imitation learning objectives.

\subsection{Multi-Price Level Strategy}
In this subsection, we introduce a novel action space specially crafted to define the multi-price level MM strategy. 

In practical MM scenarios, market participants analyze many quantities before sending orders,  among which the most important one is the distance between their target price and the "reference market price"  $p_{ref}$, typically the midprice \cite{huangQueuereactive2013}.
The LOB can be formulated as a $2K$-dimensional vector, where $K$ denotes the number of available limits on each side.
Notably, $p_{ref}$ serves as the LOB's central point, thereby determining the positions of the 2K limits $Q_{\pm i}$, 
where $Q_{\pm i}$ represents the limit at the distance $i-0.5$ ticks to the right ($+i$) or left ($-i$) of $p_{ref}$. 
 %
 It is assumed that buy limit orders are placed on the bid side, and sell ones on the ask side.
 The queue length at $Q_i$ is denoted as $l_i$.
 
Most existing MM methods adopt the market midprice as $p_{ref}$ to encode the LOB. However, the specific price linked with level $i$ may experience frequent shifts due to the dynamic fluctuations in the midprice. 
Whenever alterations occur in  $p_{ref}$, the corresponding $l_i$ instantaneously transitions to the value of one of its adjacent neighbors. 
This hinders the extraction of valid micro-market information from LOB. 
 The necessity arises to define a stable reference price.
 
 To this end, this paper formulates  $p_{ref}$ in the subsequent manner:
 Firstly, we set up a reference price $\tilde p_{ref}$ that follows the midprice. Specifically, in instances where the market spread is odd, $\tilde p_{ref,t} := m_t = \frac{ask_t+bid_t}{2}$. When the spread is even, $\tilde p_{ref,t}  := m_t\pm\frac{\text {price tick}}{2} $, with the selection of the sign based on proximity to the prior value. 
 Afterwards, at the beginning of an episode, we set $p_{ref,0} = \tilde p_{ref,0} $.  Then when the midprice $m_t$ increases (or decreases), only if $l_{-1,t}  = 0$ (or $l_{1,t} = 0$),   
 $p_{ref,t}$ is updated to $\tilde p_{ref,t} $. 
Consequently, changes of $p_{ref}$ are possibly caused by one of the three following events: 
(1) The insertion of a buy/sell limit order within the bid-ask spread while $Q_1$/$Q_{-1}$ is empty. (2) A cancellation of the last limit order at one of the best price. (3)  A market order that consumes the last limit order at one of the best offer queues. 
 Note that within our framework, the LOB accommodates empty limits, as depicted in Figure \ref{fig:pref}.
In this way, we obtain a more stable $p_{ref}$\footnote{See detailed explanations in the supplementary materials.\label{foot:supply}}, enabling  effective encoding of LOB and multi-price level orders.

\begin{figure}[t]
  \centering
  \includegraphics[width=.9\linewidth]{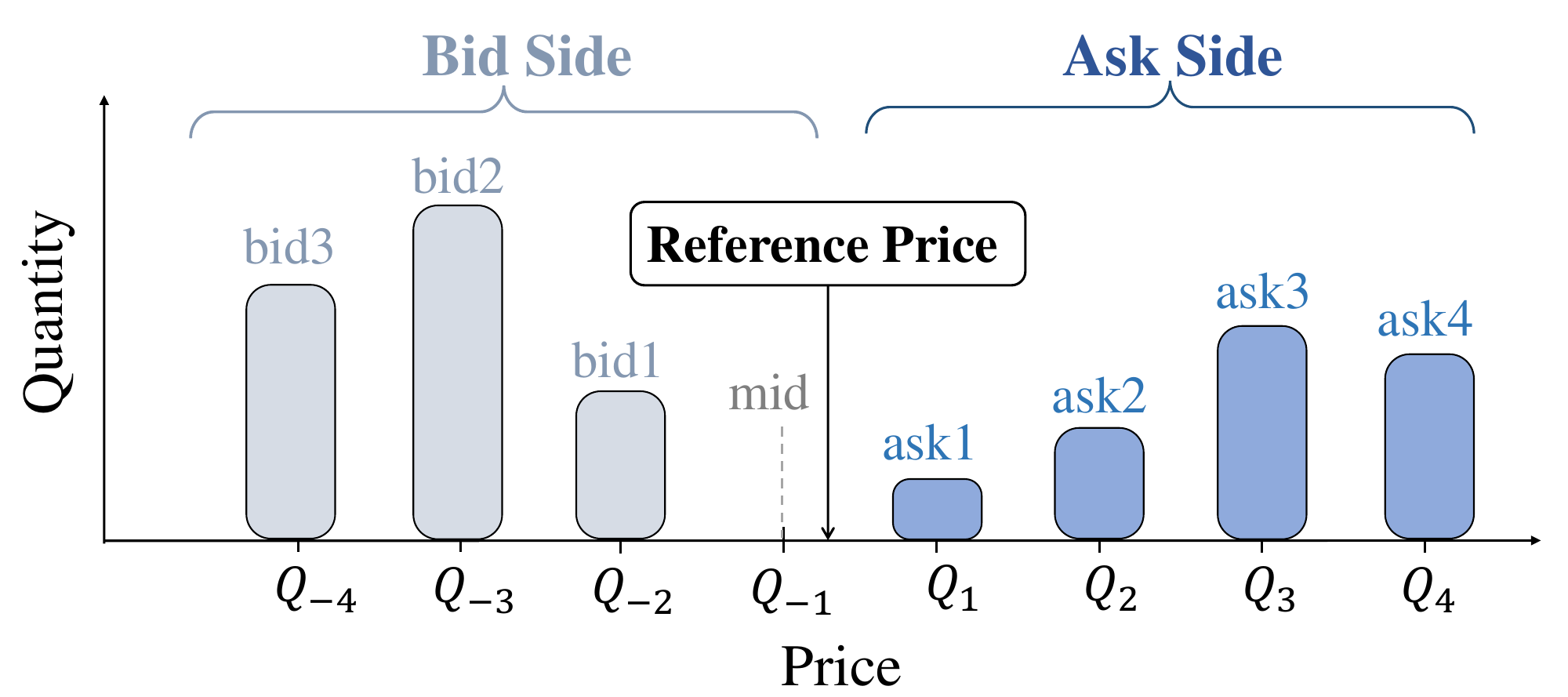}\\
  \caption{Limit order book and the reference price. In this instance, $Q_{-1}$ is an empty limit.}
  \label{fig:pref}
\end{figure}

\subsubsection{State Space}
With such a stable reference price, IMM then effectively encodes both micro-level market information and multi-price level orders. To mitigate adverse selection risk, the state space necessitates the inclusion of macro-level market information\textsuperscript{\ref{foot:supply}}.
At time step $t$, the IMM agent observes the state formulated by:
\begin{equation}
\label{eq:state}
    \bm{s}_t = (\bm{s}_t^{m},\bm{s}_t^s,\bm{s}_t^{p}),
\end{equation}
 where $\bm{s}_t^m$ denotes the \textit{market variables} encoding the current market status;  $\bm{s}_t^s$ denotes the \textit{signal variables}, including multi-granularity auxiliary predictive signals; $\bm{s}_t^p$ denotes the \textit{private variables}, including: the current inventory $z_t$, the queue position information and volume of the agent's orders that rests on the LOB, denoted by $s_t^q = (q_t^{-K}...,q_t^{-1},q_t^{1},...,q_t^K)$ and 
 $
 s_t^v = (v_t^{-K}...,v_t^{-1},v_t^{1},...,v_t^K) 
 $ respectively. 
 Here $q_i$ and $v_i$ denote the queue position information and volume at price level $i$ respectively. 
 Suppose there are $m_i$ orders resting at level $i$ placed at different time step. The queue position value of the $j$-th order at level $i$ can be defined as
$
q_t^{i,j} = \frac{l_i^{front,j}}{l_t^i},
$
 where $l_i^{front,j}$ is the queue length in front of this order.
 Thus the  queue position value at  price level $i$  can be defined as the volume weighted average of the queue position values of the $m_i$ orders:
  $
q_t^i = \sum\frac{l_t^{front,i,j}}{l_t^i}
  \frac{v_t^{i,j}}{l_t^i}.
  $
 Through covering the information of the current quotes in states, the IMM  learns to avoid frequent order cancellations and replacements.

\subsubsection{Action Space}

\begin{figure}[t]
  \centering
  \includegraphics[width=\linewidth]{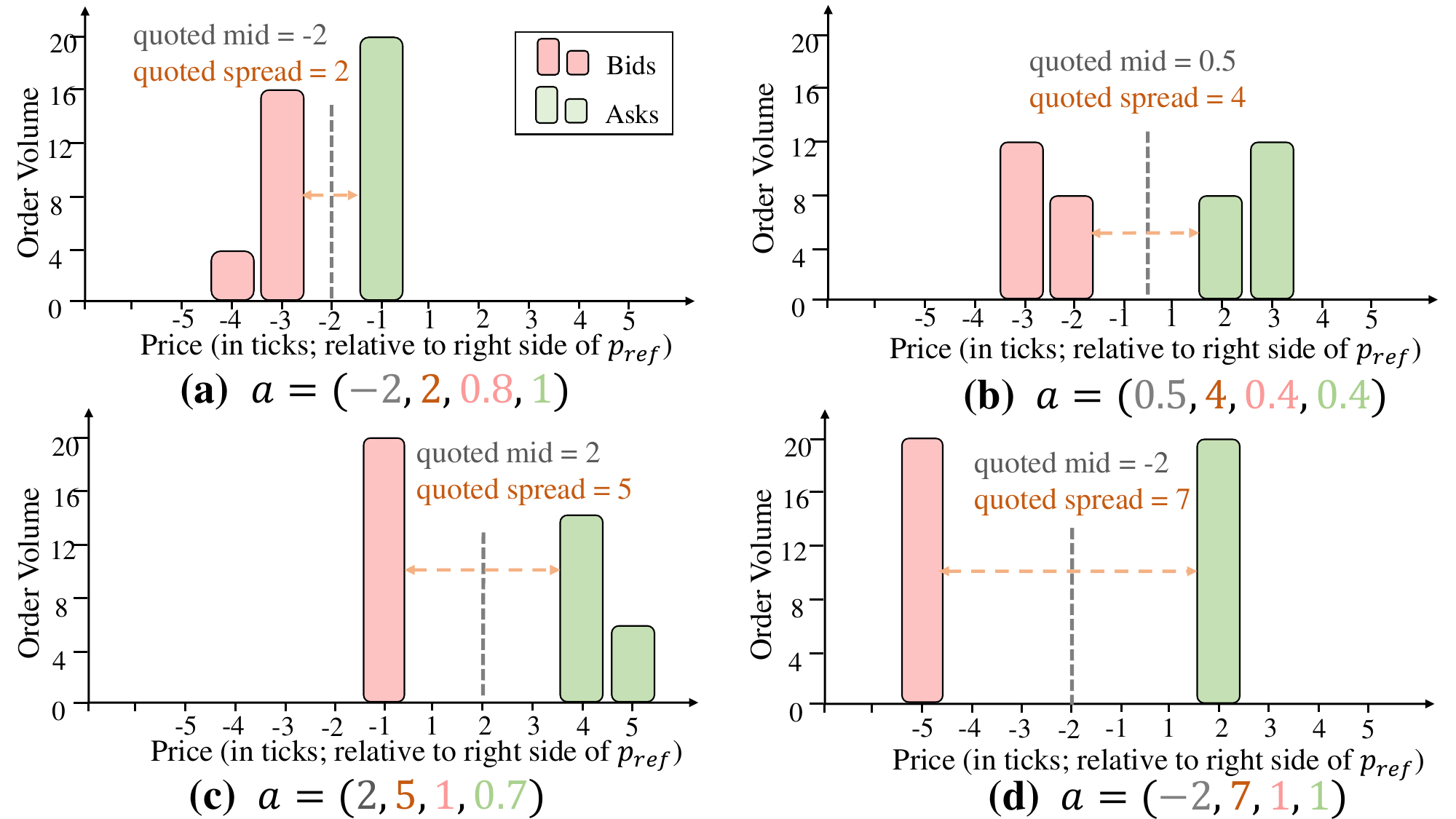}\\
  \caption{Illustration of action space. Consider a scenario where an agent positions two-level orders on both sides of the LOB,   $\phi_t^{ask}$ (the green number) determines the ratio of volume placed at the two adjacent ask price levels (heights of the green bars).}
  \label{fig:action}
\end{figure}

Reserving good queue positions beyond best bid/ask levels holds advantages of  controlling  adverse selection and non-execution risks. Therefore, practitioners tend to 
deploy order stacking strategies that 
place limit orders at multiple price levels in advance.
 IMM introduces an action encoding that expresses the complex multi-price level strategies  within a low-dimensional space.
 At time step $t$, the action $\bm{a}_t$ is defined as
\begin{equation}
\label{eq:action}
\bm{a}_t = (m_t^*, \delta_t^*, \bm\phi^{bid}_{t}, \bm\phi^{ask}_{t}),
\end{equation}
where $m_t^*$ and $\delta_t^*$ denote the desired quoted midprice w.r.t $p_{ref}$ and spread respectively. This implies that the agent's target selling price is no lower than $m_t^*+\delta_t^*/2$, and  the highest buying price is $m_t^*-\delta_t^*/2$. 
$\bm\phi^{bid}_{t}$ and $\bm\phi^{ask}_{t}$ represent parameter vectors that govern the volume distribution of the multi-level quotations. An instance of diverse two-price level actions is illustrated in Figure \ref{fig:action}.
By adopting such action formulation, the agent gains the flexibility to determine both the width  and asymmetry of the quotes with respect to the reference price.  

We formulate MM as an episodic RL task. The MM procedure allows for multi-price level order stacking, as specified below:
    (1) Choose a random start time for the episode and initialize the environment and the simulator.
    (2) Let the agent choose the desired volumes and price levels at which the agent would like to be positioned in the LOB.
    (3) Turn these desired positions into orders, including cancelling orders from levels with too much volume and placing new limit orders. 
    (4) Match the orders in market-replay simulator according to the price-time priority. (5) Update the agent’s cash and inventory of the traded asset and track profit and loss.
    (6) Repeat steps (2-4) until the episode terminates.
A picture illustration can be found in Figure \ref{fig:env} in the supplementary material.

\subsection{State Representation Learning Unit (SRLU)}
\label{sec:repr}

\subsubsection{Signal Generation }
To leverage the  labels containing future information, IMM pretrains supervised learning (SL) models to generate both short- and long-term trend signals\textsuperscript{\ref{foot:supply}}. The choice of the SL models are flexible.
 In this paper, we adopt LightGBM \cite{lightgbm2017}, a highly robust ensemble model based on decision trees, to generate four multi-granularity trend signals denoted by $(y^{20},y^{120},y^{240},y^{600})$, which are the labels of  price movement trend after $1/6,1,2,5$ minutes respectively.
 While training the RL policy, the parameters of the pre-trained predictors are frozen, and the outputs constitute the auxiliary signal variables $\bm{s}^s$.

\subsubsection{Attention-based Representation Learning}
\begin{figure}[t]
  \centering
  \includegraphics[width=\linewidth]{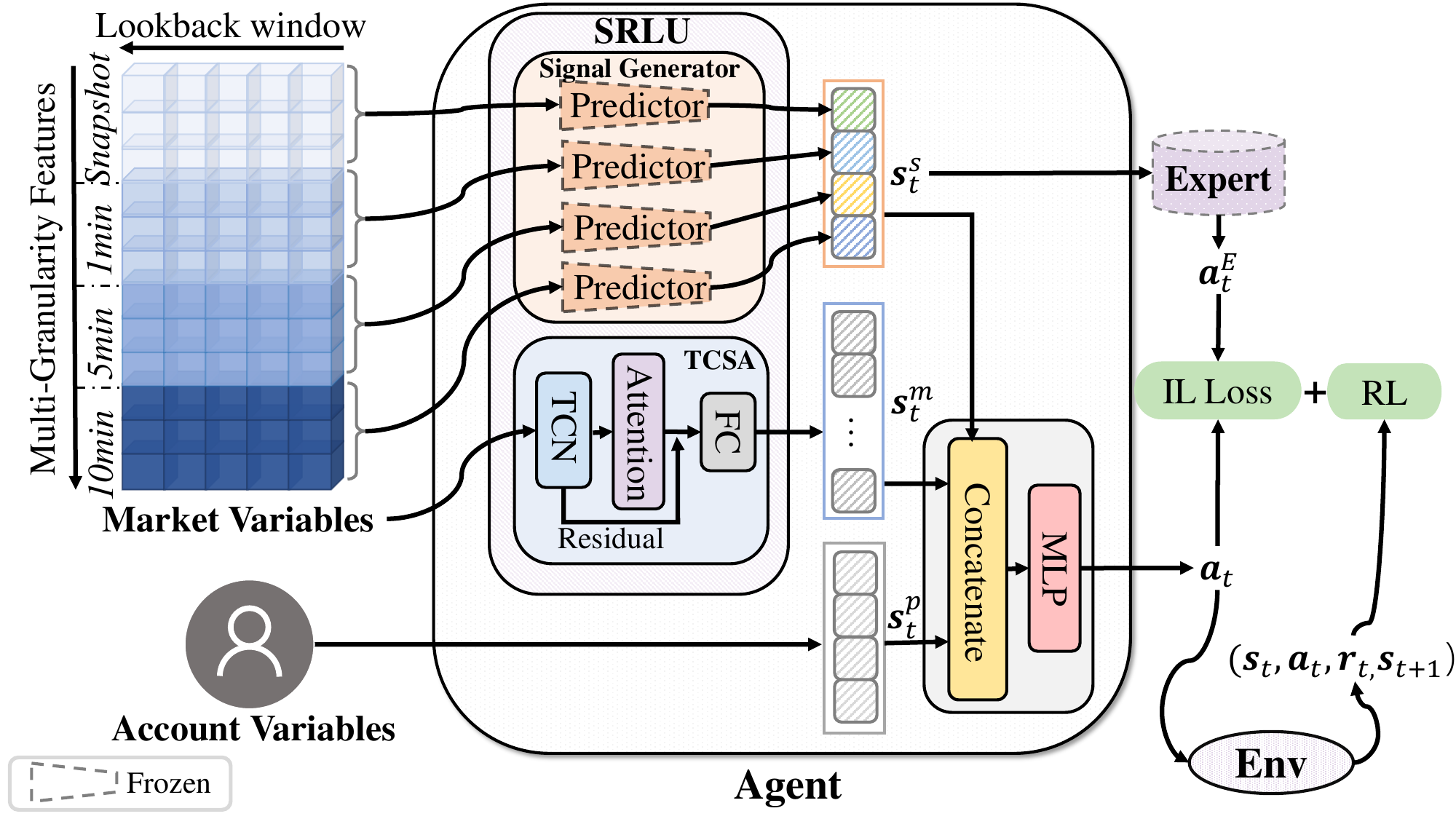}\\
  \caption{The proposed IMM learning framework.}
  \label{fig:algo}
\end{figure}

Deep RL algorithms usually suffer from the low data-efficiency issue. 
Besides the auxiliary signal prediction, we propose an temporal convolution
and spatial attention (TCSA) network to extract additional effective representations from  the noisy market data $\bm{x}$. The structure of TCSA is depicted in Figure \ref{fig:algo}.

The proposed approach IMM first utilizes a temporal convolution network (TCN) \cite{yu2015multi} block to extract the time-axis relations in the data. Compared to recurrent neural networks, TCN has several appealing properties including parallel computation and longer effective memory. 
After conducting TCN operations on $\bm{x} \in \mathbb R^{F \times L}$ along the time axis, we obtain an output tensor denoted by $\hat{\bm H} \in \mathbb R^{F \times L}$, where $F$ is the dimension of features, and $L$ is the temporal dimension.  

Afterward, IMM adopts an attention mechanism \cite{vaswani2017attention} to handle the spatial relationships among different features.  Given the output vector of TCN, we calculate the spatial attention weight as 
$
    \hat{\bm{S}} = \bm V \cdot \text{sigmoid}\big((\hat{\bm H}\bm W_1)\bm \cdot (\hat{\bm H}\bm W_2)^T + \bm b \big) ,
    $
where $\bm W_1, \bm W_2 \in \mathbb R^{L}$, and $\bm V \in \mathbb R^{F\times F}$ are parameters to learn, $\bm b \in \mathbb R^{F\times F}$ is the bias vector. The matrix $\bm \hat{S} \in \mathbb R^{F\times F}$ is then normalized by rows to represent the correlation among features:
$
    \bm S_{i,j} = \frac{\exp(\hat{\bm S}_{i,j})}{\sum_{u = 1}^F \exp(\hat{\bm S}_{i,u})}, \forall 1\leq i \leq F.
$

We adopt the ResNet \cite{he2016resnet} structure to alleviate the vanishing gradient problem in deep learning. The final representation abstracted from $\bm x$ is denoted by $H = S\times \hat{\bm H} + \bm x$, and it is then translated to a vector with dim $F'$ using a fully connected layer:
$\bm s^m = \text{sigmoid}(W_4\cdot\text{ReLU}(\bm H W_3 + \bm b_3) + \bm b_4).$
The representation $\bm s^m$ is concatenated with the signal state $\bm{s}^s$ and private state $\bm s^p$. 

\begin{table*}[t!]
    \resizebox{\textwidth}{!} 
    { 
        \begin{tabular}{l|ccc|ccc|ccc|ccc}
        
        \hline
    
          & \multicolumn{3}{c|}{RB} 
          & \multicolumn{3}{c|}{FU}  & \multicolumn{3}{c|}{CU} 
          & \multicolumn{3}{c}{AG}
          \\
        
        \hline

        & EPnL[$10^3$] &  MAP[unit] & PnLMAP  
        & EPnL[$10^3$] &  MAP[unit] & PnLMAP  
        & EPnL[$10^3$] &  MAP[unit] & PnLMAP  
        & EPnL[$10^3$] &  MAP[unit] & PnLMAP  
        \\

        \hline
        
        FOIC 
        & 3.23 $\pm$ 4.35  & 255 $\pm$ 111 & 14 $\pm$ 22 
        & -7.79 $\pm$ 9.25 & 238 $\pm$ 135 & -43 $\pm$ 56 
        & -33.05 $\pm$ 27.63  & 206 $\pm$ 141 & -161 $\pm$ 224 
        & -48.39 $\pm$ 28.83 & 189 $\pm$ 154 & -250 $\pm$ 335  
        \\
        
        LIIC  
        & 2.26 $\pm$ 3.32 & 123$\pm$32 &  20 $\pm$ 29   
        &   -6.89 $\pm$  6.66 & 115$\pm$ 30 & -66 $\pm$ 69 
        & -24.19 $\pm$ 14.83  & 150 $\pm$ 20& -164 $\pm$ 513 
        & -38.9 $\pm$ 26.2 & 142 $\pm$ 45 & -302 $\pm$ 243  
        \\

        LTIIC 
        & 9.16 $\pm$ 4.87 & 65 $\pm$ 6 & 139 $\pm$ 68 
        & 8.26 $\pm$ 2.64 & 52 $\pm$ 3 & 160 $\pm$  50 
        & -16.74 $\pm$ 15.81  & 112 $\pm$ 109 & -190 $\pm$ 203 
        & -32.57 $\pm$ 22.8 & 128 $\pm$ 22 & -264 $\pm$ 166  
        \\
        
        \hline
        $RL_{SD}$ 
        & 4.36 $\pm$ 1.64 & {\bf 38} $\pm$ {\bf 4} & 114 $\pm$ 38 
        & 7.31 $\pm$ 5.38 & 76 $\pm$ 29 & 90 $\pm$  46
        & -19.7 $\pm$ 17  & 214 $\pm$ 109 & -92 $\pm$ 298 
        & -25.43 $\pm$ 23.83 & 107 $\pm$ 37 & -237 $\pm$ 235 
        \\
        
        $DRL_{OS}$
        & 8.22 $\pm$ 3.70 & 51 $\pm$ 4 & 156 $\pm$ 61 
        & 11.03 $\pm$ {13.87} & {\bf 37} $\pm$ {\bf 3} & 30 $\pm$ 36 
        & -18.9 $\pm$ 18.02  & 647 $\pm$ 2367 & -99 $\pm$ 147 
        & -28.39 $\pm$ 27.92 & 169 $\pm$ 154 & {\bf -167 $\pm$ 135} 
        \\

        \hline
        {\bf IMM}  
        & {\bf 16.46} $\pm$ {\bf 9.10}  & 96$\pm$ 13 & {\bf 165} $\pm$ {\bf 74}  
        &{\bf 28.10} $\pm$ {\bf 10.27}  & 102  $\pm$  14 & {\bf 274} $\pm$ {\bf  89}  
        &{\bf -4.86} $\pm$ {\bf 10.17}  & {\bf 111  $\pm$  28} & {\bf-43} $\pm$ {\bf 87}  
        &{\bf -14.5} $\pm$ {\bf 20.2}  & 102  $\pm$  14 & { -274} $\pm$ {  89}  
        \\

        \hline
        \end{tabular}
        
    }
    \caption{The comparison results of the proposed method and the benchmarks.}
    \label{tab:res}
\end{table*}

\subsection{Imitative Reinforcement Learning Unit (IRLU)}
\label{sec:RL}

\subsubsection{A Signal-Based Expert   }
\label{sec:expert}
To guide efficient exploration, we define a linear suboptimal rule-based expert strategy named  Linear in Trend and Inventory with Inventory Constraints (LTIIC), which is commonly used by human experts. Readers might employ other effective expert strategies if available.
 $ LTIIC(a,b,c,d)$ corresponds to a strategy where a market maker adjusts its quote prices based on both its inventory level and trend prediction signals. 
 If $ - d\leq z_t\leq d$  at  time $t$, the ask and bid orders are placed with prices:
    \begin{equation} 
    \label{eq:ltiic}
        \left\{
        \begin{aligned}
            ask_t^q & = & m_t + a + b\cdot z_t  + c\cdot \hat y_t,\\
            bid_t^q & = & m_t - a + b\cdot z_t  + c\cdot \hat y_t,\\
        \end{aligned}
        \right.
    \end{equation}
where $a$, $b$, $c$ and $d$ are predetermined  parameters, $m_t$ stands for the midprice of the LOB, $z_t$ represents the inventory, and $\hat y_t\in\{-1,0,1\}$ signifies a short-term predictive trend signal. The insights of LTIIC strategy lies in that during a short-term upward market trend, ask-side limit orders are more likely to be executed than bid-side ones. 
A logical approach involves implementing a narrow half-spread on the bid side and a broader half-spread on the ask side, thus reducing the risk exposure due to adverse selection. Simultaneously, the trader adjusts parameter $b$ to regulate the inventory level, while $a$ determines the quoted spread.
In cases where  $z_t >= d$  or $z_t <= -d$, only orders on the side opposite to the inventory are posted. 

\subsubsection{Policy Learning }
 We utilized the actor-critic  RL framework \cite{konda1999actor}, where the critic evaluates the action taken by the actor by computing the value function, and the actor (policy) is optimized to maximize the value output by the critic. To improve the sample efficiency, we use the off-policy actor-critic method TD3 \cite{fujimoto2018addressing} as the base learner, and the policy $\pi$ is updated with the deterministic policy gradient \cite{silver2014deterministic}:
\begin{equation}
\label{eq:actor}
    \pi = \arg\max_{\pi} \mathbb E_{(\bm s, a)\sim D}\big[Q(\bm{s}, \pi(\bm{s}))\big],
\end{equation}
where $Q$ is a value function approximating the expected cumulative reward, $Q(\bm{s}_t, \bm{a}_t)=\mathbb E[\sum_{i=t}^T\gamma^{i-t}r_i|\bm{s}_t, \bm{a}_t]$.

In Equation (\ref{eq:actor}), $D$ denotes the replay buffer collected by a behavior policy, which is generated  by adding some noise to the learned policy $\pi$. Following the TD3 method \cite{fujimoto2018addressing}, the value function $Q$ is optimized in a twin delayed manner with the data sampled from both the replay buffer $D$ and expert dataset $D_E$.

Since the high-dimensional state space, the complex action space, and the stochastic trading environment induce a hard exploration problem, learning with a pure RL objective in Equation (\ref{eq:actor}) is extremely difficult. 
To promote the policy learning in such complex trading environment, we propose to augment the RL method with the objective of imitating the quoting behavior in an expert dataset $D_{E}$ as:
\begin{footnotesize}
\begin{equation}
\label{eq:bc}
\begin{aligned}
    \pi = & \arg\max_\pi \mathbb E_{(\bm{s}, a)\sim D}\big[  Q(\bm{s}, \pi(\bm{s}))\big]
    - \mathbb E_{(\bm{s}, \hat{a})\sim D_{E}}\big[\lambda \cdot (\pi(\bm{s})-\hat{a})^2\big],
    \end{aligned}
\end{equation}
\end{footnotesize}
where $\lambda$ is a scaling coefficient that balances  maximizing the Q values and minimizing the behavior cloning (BC) loss.  We set $\lambda$ decrease with the growth of the training steps.

As the expert dataset contains reasonable suboptimal MM behaviors, the agent benefits from the imitation learning techniques  through abstracting advanced trading knowledge. Thus the proposed method could achieve more efficient exploration and policy learning in the highly stochastic market environment compared to the RL methods without imitation learning.

\subsubsection{Reward Function for Diverse Utilities} 
The decision process of the market makers is subject to several trade-offs, including  probability of execution and spread, inventory risk, and compensation from the exchange. 
 To meet the diverse utilities of market makers, three factors are proposed to be considered: 
 
  \textbf{Profit and loss ($PnL$)}. PnL is a natural choice for the problem domain, comprising a realized $PnL$ term (left part) and a floating $PnL$ term (right part), given by:
    \begin{footnotesize}
        \begin{equation}
        \label{eq:pnl}
        PnL_t = \bigg(\sum_{i\in A_t}p^a_{i}\cdot v^a_i -\sum_{j\in B_t}p^b_{j}\cdot v^b_j\bigg) +(p_{t+1} - p_{t}) \cdot z_{t+1},
        \end{equation}
    \end{footnotesize}
where $p$ denotes the market midprice; $p^a,v^a,p^b,v^b$ represents the price and volume of the filled ask (bid) orders  respectively; $z$ signifies  the current inventory, with $z > 0$ when the agent holds a greater long position than short. 

  \textbf{Truncated Inventory Penalty}. 
To mitigate inventory risk, it is reasonable to introduce an additional inventory dampening term.
Considering that advanced market makers may choose to hold a non-zero inventory to exploit clear trends while capturing the spread, 
an enhanced approach involves applying the dampening term solely to higher-risk inventory levels:
    \begin{footnotesize}
    \begin{equation}
        \label{eq:penalty}
        IP_t = - \eta |z_t| \cdot \mathbb I(|z_t| > C),
    \end{equation}
    \end{footnotesize}
A penalty for inventory holding is applied solely when the inventory $z_t$ surpasses a constant $C$.

  \textbf{The Market Makers’ compensation from the exchange} constitutes a primary revenue stream for numerous market makers  \cite{2015fernadez_introlob}. Therefore, ensuring a substantial volume of transactions to secure compensations holds significant importance for a variety of MM companies. 
  To this end, a bonus term is incorporated to encourage transactions of the agent:
\begin{footnotesize}
\begin{equation}
    C_t =\beta \bigg( \sum_{i\in A_t}p^a_{i}\cdot v^a_i +\sum_{j\in B_t}p^b_{j}\cdot v^b_j\bigg),
\end{equation}
\end{footnotesize}

Ultimately, by appropriately tuning the parameters $\eta$ and $\beta$ based on personalized utilities, 
IMM ensures alignment with the requirements of a broad spectrum of market makers, employing the combination of these three categories of rewards:
\begin{footnotesize}
\begin{equation}
    \mathcal R(\bm{s}_t, \bm{a}_t, \bm{s}_{t+1}) =  PnL_t +  C_t +  IP_t.
\end{equation}
\end{footnotesize}

\section{Experiments}
\label{sec:exp}


\subsection{Experimental Setup}

We conduct experiments on four datasets comprised of historical data of the spot month contracts of the $FU$, $RB$, $CU$ and $AG$ futures from the Shanghai Futures Exchange\footnote{Here $RB$, $FU$, $CU$ and $AG$ refers to the Steel Rebar,  Fuel Oil, Copper, and Silver Futures Contracts respectively. 
}.  
The data consists of the 5-depth LOB and aggregated trades information associated with a $500$-milliseconds real-time financial period. 
We use the data from July 2021 to March 2022 (126 trading days) for training with $20\%$ as the validation set, and test model performance on April 2022 $\sim$ July 2022 (60 trading days).
 In each episode, the agent adjusts its 2-level bids and asks every $500$ milliseconds, with a fixed total volume $N = 20$ units on each side. 
The episode length is set to 1.5 trading hour, with $T=10800$ steps. 

\subsubsection{Benchmarks}
\label{seq:bench}

We compare IMM with three  rule-based benchmarks and two state-of-the-art RL-based approaches:
\begin{enumerate}
    
    \item \textbf{FOIC} represents a Fixed Offset with Inventory Constraints strategy introduced by \cite{Gaperov2021MarketMW}.  $FOIC(d)$ refers to the strategy that posts bid (ask) orders at the current best bid (ask) while adhering to the inventory constraint $d$.  
    
    \item \textbf{LIIC}. A Linear in Inventory with Inventory Constraints strategy   \cite{Gaperov2021MarketMW} corresponds to the  strategy where a market maker adjusts its quote prices based on its inventory level. The quotes of LIIC can be formulated as $LTIIC(a,b,c=0, d)$ using Equation (\ref{eq:ltiic}).

     \item \textbf{LTIIC} is the expert adopted in  IMM.

     \item { $\bf RL_{DS}$ } refers to a RL-based single-price level strategy proposed by \cite{Spooner2018MarketMV}.
    
     \item { $\bf DRL_{OS}$ }  refers to a state-of-the-art RL-based multi-price level strategy proposed in \cite{ICAIF2022os}. The agent decides whether to retain one unit of volume on each price level, not allowing for volume distribution across all price levels. 

\end{enumerate}

\subsubsection{Evaluation metrics}
We adopt four financial metrics to assess the performance of a MM strategy:
\begin{itemize}
    
    \item \textbf{Episodic PnL} is a natural choice to evaluate the profitability of a MM agent, since there is no notion of starting capital in MM procedure: 
        $EPnL_T = \sum_{t=1}^T PnL_t.$
    
    \item \textbf{Mean Absolute Position (MAP)} accounts for the inventory risk, defined as:
        $MAP_T = \frac{\sum_{t=1}^T |z_t|}{\sum_{t=1}^T 1\cdot  \mathbb I(|z_t|>0)}$.
    
    \item \textbf{Return Per Trade (RPT)}  evaluates the agent's capability of capturing the spread. It is normalized across different markets by the average market spread $\overline{\delta^m}$ .
    \begin{footnotesize}
     \begin{equation*}
         RPT_T = \bigg( \frac{\sum_{i\in A_T} p_i^a*n_i^a}{\sum_{i\in A_T} n_i^a} - 
            \frac{\sum_{j\in B_T} p_j^b*n_j^b}{\sum_{j\in B_T} n_j^b}\bigg)\bigg/ \overline{\delta^m}. 
     \end{equation*}
    \end{footnotesize}
    
    
    \item \textbf{PnL-to-MAP Ratio (PnLMAP)}  simultaneously considers the profitability and the incurred inventory risk of a market making strategy:
        $PnLMAP_T = \frac{EPnL_T}{MAP_T}$.
    
\end{itemize}

 \subsection{Comparison Results with Baselines}

For a fair comparison, we tune the hyper-parameters of these methods for the maximum PnLMAP$_T$ value  on the validation dataset.
The comparison results of  IMM and the benchmarks on the four test datasets are given in Table \ref{tab:res} and supplementary materials. 
These comparison results indicate that the proposed approach significantly outperforms the benchmarks in terms of both profitability and risk management.

\begin{figure}[t]
  \centering
  \includegraphics[width=\linewidth]{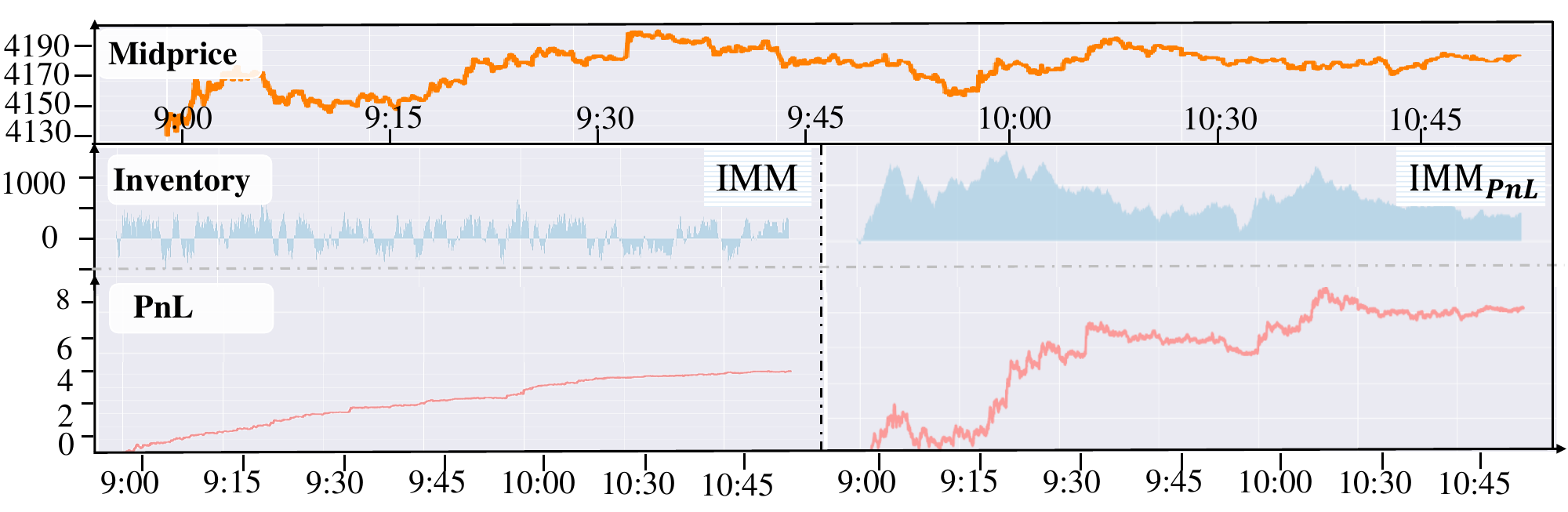}\\
  \caption{Performance of IMM and IMM$_-PnL$  on FU on Jun. 14th, 2022.}
  \label{fig:intraday}
\end{figure}

As demonstrated in Table \ref{tab:res}, on the RB dataset, the proposed method attains the highest terminal wealth as well as risk-ajusted return, albeit with a slightly elevated MAP in comparison to the expert and the two RL-based methods. Besides, the two multi-price level RL-based agents  $DRL_{OS}$ and IMM  outperform the single-price level method $RL_{DS}$,  indicating the superiority of mult-price level strategy. 
On the FU dataset, IMM not only achieves the highest terminal wealth but also demonstrates the most favorable return-to-risk performance and spread-capturing ability, while maintaining the second-lowest inventory level. The  RL-based strategies achieve commendable performance compared to the rule-based strategies which fail to make profits in most trading days.
 Moreover,  it is observed that IMM acquires a competitive MM strategy, which attains stable dividends with profits (the pink line) while sustaining the inventory at a  tolerable low level (the compact blue region) in a violate market, as illustrated in the left part of Figure  \ref{fig:intraday}. The inventory fluctuates around zero, which is a desirable behaviour. Remarkably, since IMM does not force the agent to place opposite-side orders to clear its inventory, it is a quite appealing result to see IMM accomplishes automatic inventory control based on   state-derived information. 
 Even in the  challenging  MM tasks on CU and AG markets which show lower market liquidity,
 the proposed method significantly outperforms the benchmarks in terms of the terminal wealth and risk-adjsted return.

  \begin{figure}[t]
  \centering
  \includegraphics[width=\linewidth]{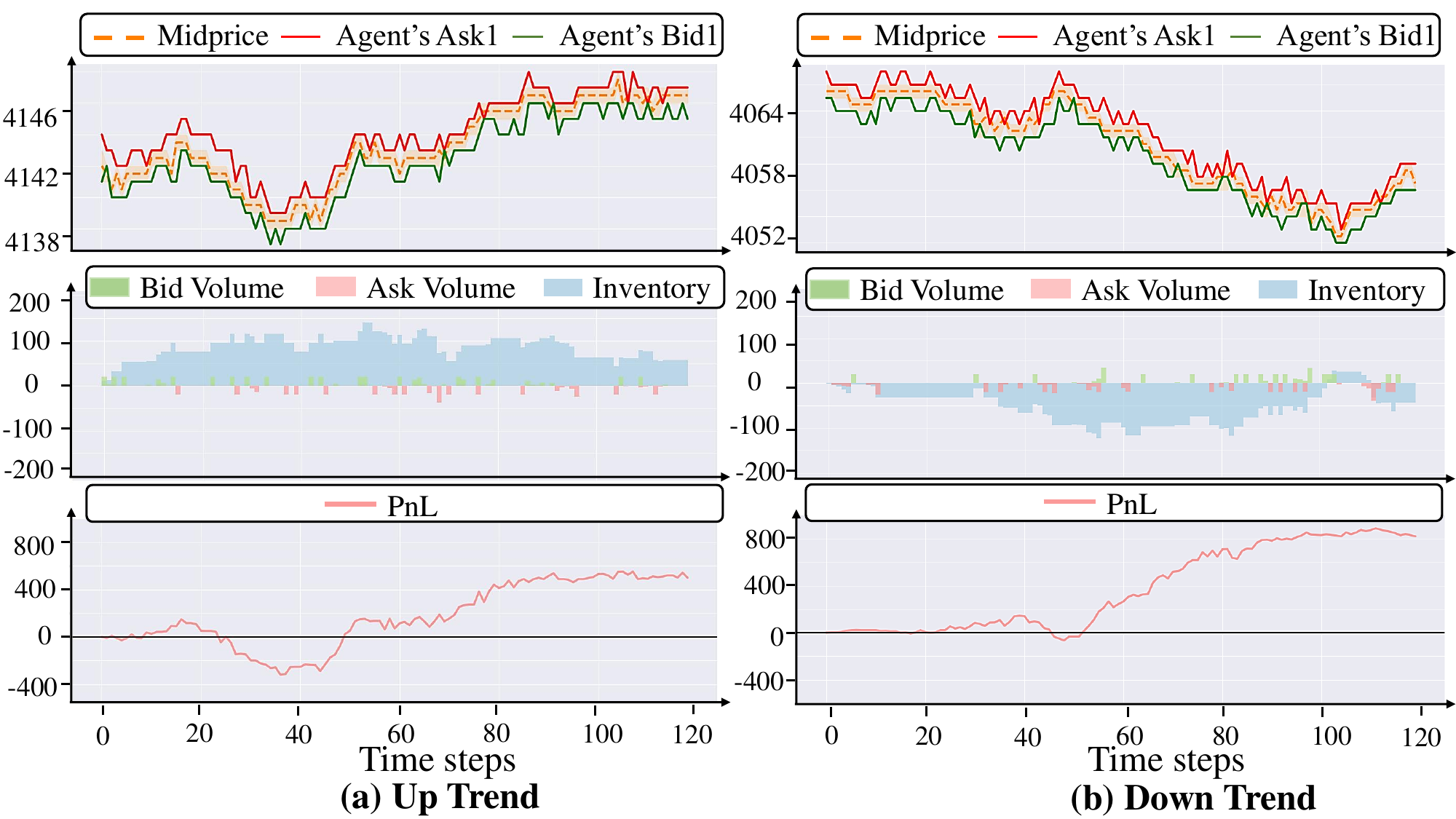}\\
  \caption{1-minute Cases. IMM performs well in both up- and down-trend markets.}
  \label{fig:trend}
\end{figure}

\subsection{Model Component Ablation Study}

To investigate the effectiveness of the model component,  we compare the proposed IMM with its five variations 
summarized in Table \ref{tab:mdl}, and the  results are listed in Table \ref{tab:abl}.

\subsubsection{Effectiveness of the State Representations}
\label{sec:abl_state}

\begin{figure}[b]
  \centering
  \includegraphics[width=.95\linewidth]{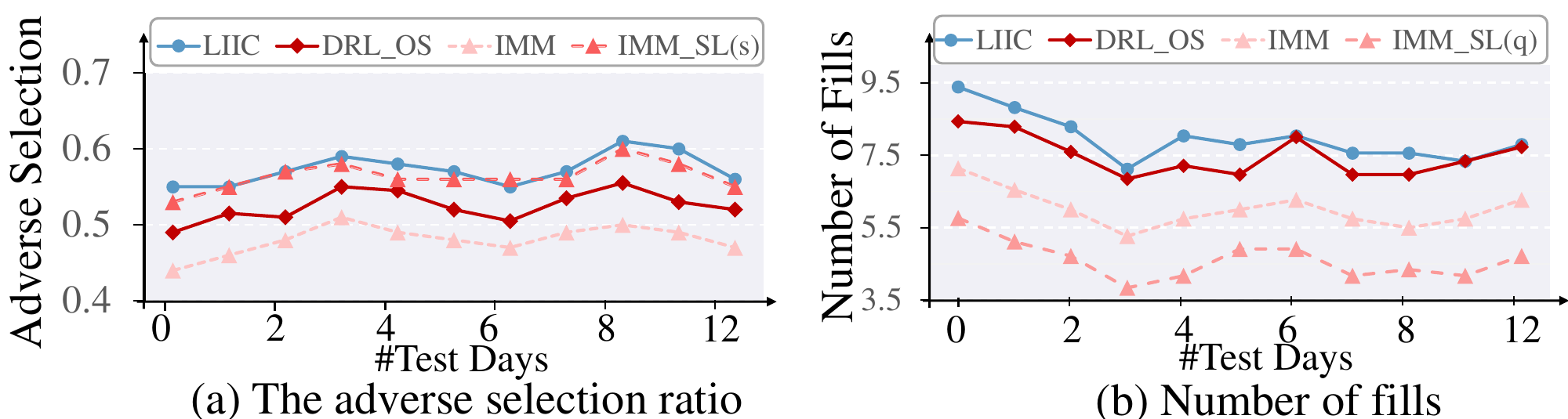}\\
  \caption{The daily adverse selection ratios and normalized number of fills.}
  \label{fig:adverse}
\end{figure}

To examine the efficacy  of the proposed state representations, we analyse the performance of three IMM$_{SL(\cdot)}$ models.
Based on Table \ref{tab:abl}, it is evident  that introducing multi-granularity signals as auxiliary observations 
holds significant importance in enhancing the MM strategy's performance.
 Figure \ref{fig:trend} visualizes the behaviour of IMM during two 1-minute periods with different trends on FU test dataset. 
 As demonstrated in Figure \ref{fig:trend}(a), benefiting from the auxiliary signals, the IMM agent anticipates an ascending price trend and proactively maintains a long position prior to the onset of a short-term bullish trend (steps 0-40).
 With the trend terminates  (step 70-120), the agent gradually reduces its inventory through placing orders with narrow ask-side half-spread  and broad bide-side one w.r.t the market midprice.  Similarly, the IMM agent demonstrates proficient behavior in downside markets as shown in \ref{fig:trend}(b).  
 Combined with the adverse select ratio depicted in Figure  \ref{fig:adverse}, it can be concluded that  IMM   has learned to mitigated adverse selection risk.
 
For a deeper investigation into the role of auxiliary signals in improving performance,  
we calculated the adverse selection ratio as 
$adv\_ratio = \frac{\text{\# adverse fills} }{\text{\# fills in last interval}}.$ 
Here adverse fills refers to limit bid (ask) orders which are executed shortly before a downward (upward) movement of the best bid (ask) price. As if the best bid price would have gone down, it might have better to wait the next bid price level \cite{ICAIF2022os}. 
 Based on Figure \ref{fig:adverse}(a), we can deduce that the multi-granularity predictive signals play a vital role in mitigating adverse selections. That might  because they provide effective information about market conditions, which enables more flexible trade-off between spread-capturing and trend-chasing.
 Moreover, Figure \ref{fig:adverse}(b) demonstrates that the information regarding multi-price level orders additionally contributes to enhancing the fills count by minimizing frequent cancellations and preserving queue positions.

\begin{table}[t]
\centering
    \resizebox{.4\textwidth}{!} 
    { 
        \begin{tabular}{l|ccccc}

        \hline
        Models & QuotesInfo & Signals & TCSA  & RL  & IL  \\

        \hline

        IMM$_{SL(m)}$
        &  O  &  O &  X   &   O  &   O\\

        IMM$_{SL(s)}$
       &  O  &  X &  O  &   O  &  X  \\

       IMM$_{SL(q)}$
       &  X  &  O &  O  &   O  &    O\\

       IMM$_{BC(0)}$
        &  O  &  O &  O  &   O  & X   \\
        
        IMM$_{BC(1)}$
       &  O  &  O &  O  &  X &    O\\

       
        \hline
        \end{tabular}
    }
    \caption{Five variations of the proposed IMM.}
    \label{tab:mdl}
\end{table}
\begin{table}[t]
\centering    
    
    \resizebox{.45\textwidth}{!} 
    { 
        \begin{tabular}{l|cccc}
        
        
        \hline
       & EPnL[$10^3$] & {  MAP}[unit] & PnLMAP 
       & SR  \\

        \hline
        IMM$_{SL(m)}$
        & 10.57 $\pm$  8.63  & 74 $\pm$ 41 & 142 $\pm$ 39 
        & 1.22\\

        IMM$_{SL(s)}$
        & 7.83 $\pm$ 3.64 & {\bf 49}  $\pm$ {\bf5} & 159 $\pm$  46 
        &  2.15\\

        IMM$_{SL(q)}$
        & 10.20 $\pm$ 9.72 & 74  $\pm$ 47 &  104  $\pm$  56 
        & 1.05 \\
        
       IMM$_{BC(0)}$
        & 14.67  $\pm$ 5.11 & 85 $\pm$ 5 & 172 $\pm$ 57 
        & {\bf 2.87}\\
        
        IMM$_{BC(1)}$
        & 8.22 $\pm$ 3.70 & 51 $\pm$ 4 & 156 $\pm$ 61 
        & 2.22\\ 
        
        IMM   
        &{\bf 28.097}  $\pm$  {\bf 10.27}  & 103  $\pm$   {15} & {\bf 274}  $\pm$  {\bf  89}  
        &  2.80   \\
        \hline
        \end{tabular}
    }
    \caption{ Comparison results of ablation study on FU dataset.}
    \label{tab:abl}
\end{table}

\subsubsection{Effectiveness of the IRLU}
The comparison results between IMM$_{BC(\cdot)}$ and IMM 
provide empirical evidence of the importance of extracting additional knowledge from the expert and conducting efficient exploration, particularly for challenging financial tasks.
Besides, IMM outperforms the expert LTIIC strategy a lot.
 As the RL agent faces challenges in identifying a viable trading approach and sustaining it over multiple steps during the initial training phase.
 Training while pursuing the imitation learning objective facilitates the agent  in obtaining favorable rewards and drawing valuable lessons from these experiences.

\subsubsection{Effects of Reward Functions}

To investigate whether the proposed reward function meets different utilities, we train IMM  that  with three different rewards: The IMM$_{PNL}$ mehtod  trains IMM using the PnL reward ($\eta,\beta = 0$); The IMM$_{PNL+C}$ method trains IMM with the combination of the PNL and compensation reward ($\eta = 0,\beta>0$);  The IMM$_{PNL+IP}$ method trains IMM with the combination of the PNL and truncated inventory penalty reward ($\eta > 0,\beta =0$). 
  The hyperparameters resulting in the maximum PnLMAP value on the validation dataset are chosen for these models. 
 The results on the FU dataset are outlined in Table \ref{tab:rwd}.
Here the metric \#T signifies the  number of fills normalized by the episode length.


\begin{table}[htbp]
    \resizebox{.47\textwidth}{!} 
    { 
        \begin{tabular}{l|cccc}
        
        \hline
        
       & EPnL[$10^3$] & MAP[unit] & PnLMAP 
       & \#T  \\

        \hline

        IMM$_{PnL}$ 
        & 58.76 $\pm$  94.43  & 2156 $\pm$ 655 & 31 $\pm$ 48 
        & 4.43 $\pm$ 0.94\\
        
       IMM$_{PnL+C}$ 
        & 42.86  $\pm$ 123.04 & 2041 $\pm$ 465 & 27  $\pm$   68  
        &  { 4.85} $\pm$  { 1.09} \\
        
        IMM$_{PnL+IP}$ 
        & {\bf 73.07}   $\pm$   {\bf 53.83} & 756  $\pm$  289 & 90 $\pm$ 46  
        &  4.42  $\pm$  0.96 \\
        
         IMM   
        &{ 28.097}  $\pm$  { 10.27}  & {\bf 103}   $\pm$   {\bf 15} & {\bf 274}  $\pm$  {\bf  89}  
        &  {\bf 5.15}  $\pm$  {\bf 1.19}\\
         \hline
        \end{tabular}
    }

    \caption{Performance of IMM variations trained with different reward preferemces on FU dataset.}
    \label{tab:rwd}
\end{table}
As shown in Table  \ref{tab:rwd}, the IMM$_{PNL}$ strategy tends to have the most substantial inventory risk exposure. 
We depict an example of the intra-day performance of the IMM$_{PNL}$ policy in the right part of Figure \ref{fig:intraday}.
We observe that the IMM$_{PNL}$ agent learns to chase trends through maintaining a large inventory ($>1000$). 
This result in poor out-of-sample performance with large variance. Therefore, the truncated inventory penalty term proves crucial in curbing  blind trend-chasing tendencies.

 The IMM$_{PnL+C}$ strategy also grapples with elevated inventory risk, yet it achieves a greater number of transactions  \#T compared to IMM$_{PnL}$. 
 The IMM$_{PnL+IP}$ strategy  attains the highest average terminal wealth and the return per trade metric, but concurrently records the lowest \#T, a circumstance that may be less advantageous for risk-averse market makers.
 The strategy trained with the proposed reward significantly improve the return-to-risk performance with the lowest MAPs, as well as a larger \#T, compared to the IMM$_{PnL+IP}$ strategies. 
 Althoug having the lowest average terminal wealth, the proposed IMM strategy acts very stably and might be the most favorable  policy among these four policies for a risk-averse market maker. Besides, note that the proposed strategy has a largest \#T, it could receive more compensation from the exchange.

\section{Conclusion}
In this paper, we propose IMM, a novel RL-based approach aimed at efficiently learning multi-price level MM policies.
IMM first introduces efficient state and action representations.
Subsequently, it pre-train a SL-based prediction model to generate multiple trend signals as effective auxiliary observations. Futhermore, IMM utilizes a TCSA network to handle the temporal and spatial relationships in noisy financial data. 
Through abstracting trading knowledge from a sub-optimal expert meanwhile interacting with the environments, IMM explores the state and action spaces efficiently. 
Experiments on four futures markets demonstrate that IMM outperforms the benchmarks, and further ablation studies verify the effectiveness of the components in the proposed method.

\clearpage

\bibliography{aaai24}

\clearpage
\end{document}